\documentclass[]{article} 
\usepackage[preprint]{colm2026_conference}

\usepackage{microtype}
\usepackage{hyperref}
\usepackage{url}
\usepackage{booktabs}
\usepackage{float}

\usepackage{lineno}

\definecolor{ourpurple}{HTML}{7760FB}
\definecolor{ourgreen}{HTML}{5CAD70}

\definecolor{darkblue}{rgb}{0, 0, 0.5}
\hypersetup{colorlinks=true, citecolor=ourgreen, linkcolor=ourgreen, urlcolor=ourgreen}

\usepackage{amsmath}
\usepackage{cleveref}
\usepackage{caption}
\usepackage{subcaption}
\usepackage{graphicx}
\usepackage{inconsolata}
\usepackage{underscore}

\usepackage[most]{tcolorbox}
\usepackage{xcolor}
\usepackage{tikz}
\usepackage{ragged2e}
\usepackage{parskip}

\definecolor{userbg}{HTML}{EAF3FF}
\definecolor{userborder}{HTML}{3B82F6}

\definecolor{aibg}{HTML}{F3F4F6}
\definecolor{aiborder}{HTML}{6B7280}

\definecolor{annred}{HTML}{FDECEC}
\definecolor{annredborder}{HTML}{D9485F}

\definecolor{annblue}{HTML}{EAF4FF}
\definecolor{annblueborder}{HTML}{2563EB}

\definecolor{anngreen}{HTML}{ECFDF3}
\definecolor{anngreenborder}{HTML}{059669}

\definecolor{textgray}{HTML}{374151}

\newtcbox{\annotbox}[4][]{%
  on line,
  enhanced,
  boxrule=0.6pt,
  colback=#1,
  colframe=#2,
  arc=2mm,
  left=1.2mm,
  right=1.2mm,
  top=0.7mm,
  bottom=0.7mm,
  boxsep=0.4mm,
  before upper={%
    \tikz[baseline=(X.base)] \node (X) [inner sep=0pt, outer sep=0pt] {};%
  },
  overlay={%
    \node[
      fill=#2,
      text=white,
      rounded corners=1mm,
      font=\bfseries\scriptsize,
      inner xsep=1.2mm,
      inner ysep=0.4mm,
      anchor=south west
    ] at ([xshift=0.6mm,yshift=0.8mm]frame.north west) {#3};
  }
}

\newtcolorbox{usermsg}{
  enhanced,
  breakable,
  width=0.85\textwidth,
  colback=userbg,
  colframe=userborder,
  boxrule=0.8pt,
  arc=3mm,
  left=2mm,right=2mm,top=1.5mm,bottom=1.5mm,
  fonttitle=\bfseries\scriptsize,
  title=User,
  attach boxed title to top left={xshift=2mm,yshift*=-2mm},
  boxed title style={
    colback=userborder,
    colframe=userborder,
    arc=2mm,
    left=1mm,right=1mm,top=0.6mm,bottom=0.6mm
  }
}

\newtcolorbox{aimsg}{
  enhanced,
  breakable,
  width=0.85\textwidth,
  colback=aibg,
  colframe=aiborder,
  boxrule=0.8pt,
  arc=3mm,
  left=2mm,right=2mm,top=1.5mm,bottom=1.5mm,
  fonttitle=\bfseries\scriptsize,
  title=AI,
  attach boxed title to top left={xshift=2mm,yshift*=-2mm},
  boxed title style={
    colback=aiborder,
    colframe=aiborder,
    coltext=white,
    arc=2mm,
    left=1mm,right=1mm,top=0.6mm,bottom=0.6mm
  }
}

\usepackage{enumitem}
\setlist[enumerate,itemize]{topsep=0pt,itemsep=0pt,leftmargin=18pt}

\newcommand{\signal}[1]{\mbox{\texttt{#1}}}
\newcommand{\domain}[1]{\mbox{\texttt{#1}}}
\newcommand{\failtag}[1]{\mbox{\texttt{#1}}}

\newcommand{\arch}[1]{{\color{ourpurple}{#1}}}
\newcommand{\stat}[1]{#1}

\newcommand{\ourdataset}{\mbox{Future-2K}}

\title{Invisible Failures in Human--AI Interactions}

\author{Christopher Potts$^{1,2}$ and Moritz Sudhof$^{1}$ \\
${}^{1}$Bigspin AI, ${}^{2}$Stanford University\\
\texttt{\{cgpotts,moritz\}@bigspin.ai} \\
}

\begin{document}

\ifcolmsubmission
\linenumbers
\fi

\maketitle

\begin{abstract}
AI systems fail silently far more often than they fail visibly. In an analysis of \stat{100K} human--AI interactions from the WildChat dataset, we find that \stat{79\%} of AI failures are \emph{invisible}: something went wrong but the user gave no overt indication that there was a problem. These invisible failures cluster into \stat{eight} archetypes that help us characterize where and how AI systems are failing to meet users' needs. In addition, the archetypes show systematic co-occurrence patterns indicating higher-level failure types. To address the question of whether these archetypes will remain relevant as AI systems become more capable, we also created and annotated a counterfactual dataset in which WildChat's 2024-era responses are replaced by those from three present-day frontier LMs. This analysis indicates that failure rates have dropped substantially, but that the vast majority of failures remain invisible in our sense, and the distribution of failure archetypes seems stable. Finally, we illustrate how the archetypes help us to identify systematic and variable AI limitations across different usage domains. Overall, we argue that our invisible failure taxonomy can be a key component in reliable failure monitoring for product developers, scientists, and policy makers.\footnote{Code/data: \url{https://github.com/bigspinai/bigspin-invisible-failure-archetypes}}
\end{abstract}

\section{Introduction}

Usage of AI services like OpenAI's ChatGPT, Anthropic's Claude, and Google's Gemini has skyrocketed. For instance, \citet{chatterji2025chatgpt} report that ChatGPT was serving 2.6B messages per day in June 2025, up from 451M in June 2024. The other major services seem to be experiencing similar growth rates \citep{views4you2025aitools}, and there is every reason to believe that these trends will continue upwards for the foreseeable future. The effects of these technological changes are now being felt at all levels of society, from individual psychology \citep{Dergaa2024,fang2025aihumanbehaviorsshape,gerlich2025ai} to national employment \citep{eloundou2024,brynjolfsson2025canaries,Brynjolfsson2025,machovec2025incorporating,weilnhammer2026}.

What is happening inside these interactions between humans and AIs?\footnote{We anticipate that, in the near future, the majority of these interactions will be between AIs, with no direct human involvement. However, our focus in this work is on human--AI interactions, so we will set this likely development aside.} This is a significant question for teams building AI products, scientists seeking to understand and improve the core technologies, and policy makers hoping to productively manage the changes AI will bring about. However, at present, we have only a very partial view into AI interactions. In industry, product teams generally monitor completion rates, response times, user satisfaction scores, and other high-level statistics, but these have to be supplemented with extensive manual review of individual cases. AI researchers have provided deep insights into specific phenomena (e.g., sycophancy, user expertise), but have not, to our knowledge, sought to offer a comprehensive framework for understanding human--AI interactions.

In this paper, we begin to develop one major component of such a framework. Our focus is on \textbf{invisible failures} in human--AI interactions: instances in which something went wrong but the user gave no overt indication that there was a problem. Examples of invisible failures include wrong answers delivered with complete confidence, outputs that look professional while missing the user's goal entirely, and conversations that end cleanly with the user holding incorrect information.

Our study is based in the WildChat dataset \citep{zhao2024wildchat}, a collection of over 1M \mbox{ChatGPT} conversations. For WildChat, users were given free access to ChatGPT in exchange for having their deidentified conversations released publicly. WildChat is the largest naturalistic conversational AI dataset available to date. We annotated \stat{100K} English-language WildChat transcripts with \mbox{Claude Opus 4.6}, using an annotation protocol that we developed and validated by having two frontier LLM annotators (\mbox{Claude Opus 4.6} and \mbox{GPT-5.4}) iteratively refine their guidelines with human oversight from us (\cref{sec:annprocess}). The resulting annotations have high Cohen $\kappa$ values and extremely high overall agreement rates (\cref{sec:annprocess}), which positions us well to use the annotations to study large-scale patterns.

Our annotations include high-level tags for whether the human--AI interaction failed or not and, if it did, the status of the failure as \failtag{visible}, \failtag{invisible}, or \failtag{mixed}. These tags reveal that the standard quality signals for human--AI interactions are woefully inadequate; of the failure cases, \stat{79\%} are invisible and another \stat{9\%} are \failtag{mixed}. In other words, the vast majority of failures are precisely the ones most likely to escape detection (\cref{sec:invis}).

The focus of our annotation effort is a taxonomy of \stat{eight} invisible failure archetypes: \stat{
    \arch{The confidence trap},    
    \arch{The silent mismatch}, 
    \arch{The drift},
    \arch{The death spiral},
    \arch{The contradiction unravel}, 
    \arch{The walkaway},
    \arch{The partial recovery},
    and
    \arch{The mystery failure}}. 
We find that this taxonomy is quite comprehensive, with over \stat{99\%} of invisible failures tagged with one of the first seven archetypes and well under \stat{1\%} tagged with \arch{The mystery failure} (indicating that we don't have a read on the nature of the failure). This suggests that the taxonomy can be a foundational component in reliable failure monitoring. The archetypes also show robust co-occurrence patterns that can provide richer insights. For example, \arch{The confidence trap} and \arch{The contradiction unravel} are strongly associated and jointly identify cases where the AI confidently contradicts itself. \arch{The walkaway} is a pervasive archetype that flags an abrupt and unexplained end to the interaction.

The WildChat transcripts derive from GPT-3.5-Turbo and GPT-4, and are more than two years old. A natural question is whether our findings are  artifacts of  outmoded models. To address this (\cref{sec:future}), we sampled \stat{2K} single-turn transcripts from our larger \stat{100K} sample and had \stat{four} present-day frontier models -- \stat{Claude Sonnet 4.6, Claude Opus 4.6, GPT-4.1, and GPT-5.4} -- provide new responses. The resulting dataset, \ourdataset, is an approximate update to WildChat. Analysis of this dataset shows that failure rates have come down considerably. However, the vast majority of failures continue to be invisible in our sense, and the archetype distribution for \ourdataset\ is very similar to that of the original WildChat, which reinforces our argument that the archetypes are an asset for system monitoring. 

Finally, to illustrate the power of these invisible failure archetypes, we show that they help us identify systematic AI limitations across different usage domains (\cref{sec:domains}). We find, for example, that \arch{The confidence trap} and \arch{The contradiction unravel} are associated with domains involving factual knowledge and objective capabilities, where conflicting statements are more likely to be apparent to users. Such patterns can help product designers establish expectations for how AI will behave in their domains, and they can help scientists and engineers identify domains requiring more attention during model development.

\section{Related work}

There are a variety of modern human--AI interaction datasets. However, they tend to be focused on specific tasks \citep{lin2025wildbench,alpaca_eval,dubois2024length} or specific model development scenarios \citep{bai2022training,ji2025pku}, or involve explicitly evaluative or adversarial dynamics \citep{zheng2023judging,chiang2024chatbotarena}. Our goal is to study more naturalistic interactions in a completely open fashion, making WildChat a natural choice \citep{zhao2024wildchat}. ShareChat \citep{yan2026sharechat}, which was released late in our own project work, could now be used to supplement WildChat, though it runs only through October 2025 and thus does not contain data from \mbox{Opus 4.6} or \mbox{GPT-5.4}, which points to an ongoing need for simulations of the sort we use in \cref{sec:future}.

There have been a number of attempts to develop frameworks for understanding and annotating dialogue and human--AI interactions
\citep{Walker1997,Hajdinjak2006,higashinaka-etal-2016-dialogue,Deriu2020,Borsci2022,kopf2023openassistant,moller2025,singh2025}. These mostly focus on very high-level signals. Researchers have complemented these efforts with study of specific phenomena, including 
sycophancy \citep{malmqvist2024sycophancy,sharma2025sycophancy,openai2025sycophancy,Chen:2025aa,hong-etal-2025-measuring,cheng2025sycophancy,cheng2026elephant}, %
confidence \citep{mielke2022reducing,jiang2021can,zhou-etal-2024-relying}, %
user expertise \citep{Huang2024,gillespie2025trust,anthropic2026fluency}, 
proactive engagement \citep{kaur2026knowing}, and %
social bias \citep{blodgett-etal-2020-language,santurkar2023opinionslanguagemodelsreflect,Gallegos2024,hu2025bias}. All of these phenomena relate to human--AI interaction failures in complex ways and can thus complement our work.

A growing literature seeks to develop best practices for using AI agents in the context of data annotation \citep{Gilardi2023,jung2024trust,li2024judges,tan-etal-2024-large,tseng2024expert,tseng2025evaluating,bojic2025}. The overarching goal is to make effective use of human expertise and AI expertise, while carrying forward best practices for data annotation in general \citep{Ide:Pustejovsky:2017}. Our own annotation effort seeks to capitalize on this literature through iterative human--AI and AI--AI development of an annotation protocol, inspired especially by the multi-agent peer discussion approach of \citet{tseng2025evaluating}.

\section{Data and methods}\label{sec:methods}

Our analysis is based in the WildChat dataset \citep{zhao2024wildchat}. WildChat captures unfiltered, unsolicited conversations between users and ChatGPT (with \mbox{GPT-3.5-Turbo} and \mbox{GPT-4} as the primary LLMs) across the full range of how people actually use AI, from casual questions to complex multi-turn professional workflows. WildChat has a total of 1,039,785 transcripts, making it the largest publicly available dataset of naturalistic conversational AI transcripts. The transcripts were collected in the period April 9, 2023, to May 1, 2024.

\subsection{Cohort selection}\label{sec:cohort}

For our analysis, we began with the subset of \stat{478,498} English-language transcripts. We then excluded (1)~conversations that produced an immediate refusal from the AI with no substantive exchange to analyze (generally due to invalid input to the LLM), and (2)~conversations tagged in the WildChat metadata as adversarial, explicit, or unclassifiable (primarily jailbreak attempts, NSFW requests, and inputs too ambiguous to categorize; see \citealt{zhao2024wildchat} for analysis of these transcripts). From the remaining examples, we randomly sampled \stat{100,000} for annotation and analysis.

\subsection{Annotation process}\label{sec:annprocess}

We began our annotation process with extensive manual review of WildChat transcripts, guided by heuristic labels assigned by Claude Sonnet 4.5 (\texttt{claude-sonnet-4-20250514}). We explored the data separately, and then we sampled 100 transcripts for manual review by each of us, to help us converge on the core failure modes and how to define them. This exploratory, data-driven process led us a taxonomy of \stat{eight} invisible failure archetypes: %
\stat{%
    \arch{The confidence trap},    
    \arch{The silent mismatch}, 
    \arch{The drift},
    \arch{The death spiral},
    \arch{The contradiction unravel}, 
    \arch{The walkaway},
    \arch{The partial recovery},
    and
    \arch{The mystery failure}}. %
Our full definitions of these archetypes, as used in the prompts for our final annotation runs, are given in \cref{appendix:archdefs}.

Our manual review clarified the role of automated annotation in this kind of analysis. We were confronted with transcripts in which users pose extremely challenging analytical questions, ask for help debugging complex computer code, switch between multiple languages, and presuppose deep knowledge of pop culture. Thorough human review of any single transcript can require extensive research and is often feasible only with AI assistance.

We therefore cast ourselves as supervisors of an annotation project involving a team of two AI annotators: \mbox{Claude Opus 4.6} (\texttt{claude-opus-4-6}) and  \mbox{GPT-5.4} (\texttt{gpt-5.4-2026-03-05}). Using 1K transcripts from our sample, these agents iteratively annotated transcripts using our signal taxonomy, compared their areas of agreement and disagreement, and refined their approaches to clarify the core concepts and achieve greater consensus. We studied the outcomes of these interactions and then instructed the agents to try specific annotation strategies. This was functionally like a standard expert annotation project, in which annotators work together to develop a core set of guidelines \citep{Ide:Pustejovsky:2017}, but it was conducted at a speed and scale that is only feasible with AI agents \citep{tseng2025evaluating}.

In the end, our annotators performed two tasks. (1)~Tagging each transcript for \textbf{basic failure} mode, i.e., whether it contains a failure and, if they does, whether the failure is \failtag{visible}, \failtag{invisible}, or \failtag{mixed}; \cref{appendix:failuremode} gives the definitions of the categories used by our annotators. 
(2)~Tagging \failtag{invisible} and \failtag{mixed} transcripts with zero or more \textbf{invisible failure archetypes} (full archetype definitions in \cref{appendix:archdefs}).

\begin{table}[tp]
    \centering
    \setlength{\tabcolsep}{1.5pt}
    \begin{subtable}{0.48\textwidth}
    \centering
    \begin{tabular}{@{} l c c c @{}}
    \toprule
     & $\kappa$ & Macro-$\kappa$ & Agree. \\
     \midrule
     Signals-only   & \textbf{0.84} & \textbf{0.74} & \textbf{0.94}  \\
     Transcript-only & 0.62 & 0.46 & 0.86 \\  
     Signal+Transcript & 0.77 & 0.63 & 0.91 \\
     \bottomrule     
    \end{tabular}
    \caption{Failure annotation agreement.}
    \label{tab:annfailmodes}
    \end{subtable}
    \hfill
    \begin{subtable}{0.48\textwidth}
    \centering
    \begin{tabular}{@{} l c c c @{}}
    \toprule
     & Micro-$\kappa$ & Macro-$\kappa$ & Agree. \\
     \midrule
     Signals-only   & \textbf{0.81} & \textbf{0.60} & \textbf{0.94}  \\
     Transcript-only & 0.52 & 0.37 & 0.84 \\  
     Signal+Transcript & 0.68 & 0.52 & 0.90 \\
     \bottomrule     
    \end{tabular}
    \caption{Archetype annotation agreement.}
    \label{tab:annarchetypes}
    \end{subtable}       
    \caption{Annotation project results, based on a sample of 10K transcripts. The annotators are \mbox{Opus 4.6} and \mbox{GPT-5.4}. Micro-$\kappa$ gives the global Cohen $\kappa$ across all categories, Macro-$\kappa$ is the mean of all the category-level $\kappa$ values, and Agree.\ is the overall agreement rate between the two annotators. \Cref{appendix:annotations} gives complete per-category breakdowns. We chose the clearly superior `Signals-only' approach, for which the annotators disagree on only \stat{6\%} of cases.}
    \label{tab:annproject}
\end{table}

We explored three primary approaches to these tasks, as summarized in \cref{tab:annproject}. On our initial, naive attempt, the annotators  made predictions directly in terms of the raw transcripts. This approach never led to satisfactory agreement levels, as seen in the middle rows in \cref{tab:annproject}. To address this, we moved to a two-stage approach. In stage~1, the annotator independently tagged transcripts at the turn level using a set of \stat{50} AI-oriented signals (e.g., \signal{ai_acknowledges_correction}, \signal{ai_implicit_refusal}) and \stat{13} user-oriented signal (e.g., \signal{user_expresses_frustration}, \signal{user_positive_feedback}). In stage~2, the annotator was given only a short textual report of the signal tagging done by Opus and \mbox{GPT-5.4}, as well as \mbox{Claude Sonnet 4.6} (\texttt{claude-sonnet-4-6}) to bring additional diversity. The signal report indicates which signals have full agreement and which have mixed agreement.\footnote{The signal annotations are a stepping stone to the archetype annotations, and so we do not focus directly on agreement levels for these categories. However, a complete report on these annotations is given in \cref{appendix:signalann}.} 

The two-stage `Signals-only' approach led to the highest agreement rates by far. Including the Transcript as well generally lowered agreement rates, as we see in the bottom rows in \cref{tab:annarchetypes}. For `Signals-only', the $\kappa$ values are ones that would typically be described as showing `moderate' to `substantial' agreement, indicating that there are still differences between the annotators. However, the overall agreement rates are extremely high. While these do not correct for chance agreement the way $\kappa$ values do, they show that the annotators have essentially the same overall behavior for our data. Since our goal is to study broad patterns rather than individual transcripts, high agreement is our primary concern.

We acknowledge that the success of this annotation effort may trace partly to the fact that \mbox{Opus 4.6} and \mbox{GPT-5.4} (and \mbox{Sonnet 4.6}) are much more capable than the models present in WildChat. This capability gap poses a challenge for real-time monitoring of frontier models, but it also creates an analytical opportunity for us: because the annotation model sits well above the capability frontier of the models used for the transcripts it analyzes, it can reliably detect failures.

Our goal in what follows is to use our annotations to study the failure modes in WildChat at scale, and to try to estimate how these failures will emerge in more capable models. For this work, we adopt \mbox{Opus 4.6} as our annotator, since it is the best model at this task in our informal estimation.

\section{Invisible failures}\label{sec:invis}

We now explore the empirical distribution of failures and failure archetypes in our dataset. \mbox{Opus 4.6} is our annotator, and it uses the `Signals-only' protocol. Our goal for this section is to begin to understand how and why human--AI interactions fail.

\begin{figure}[tp] 
    \centering
    \includegraphics[width=1\linewidth]{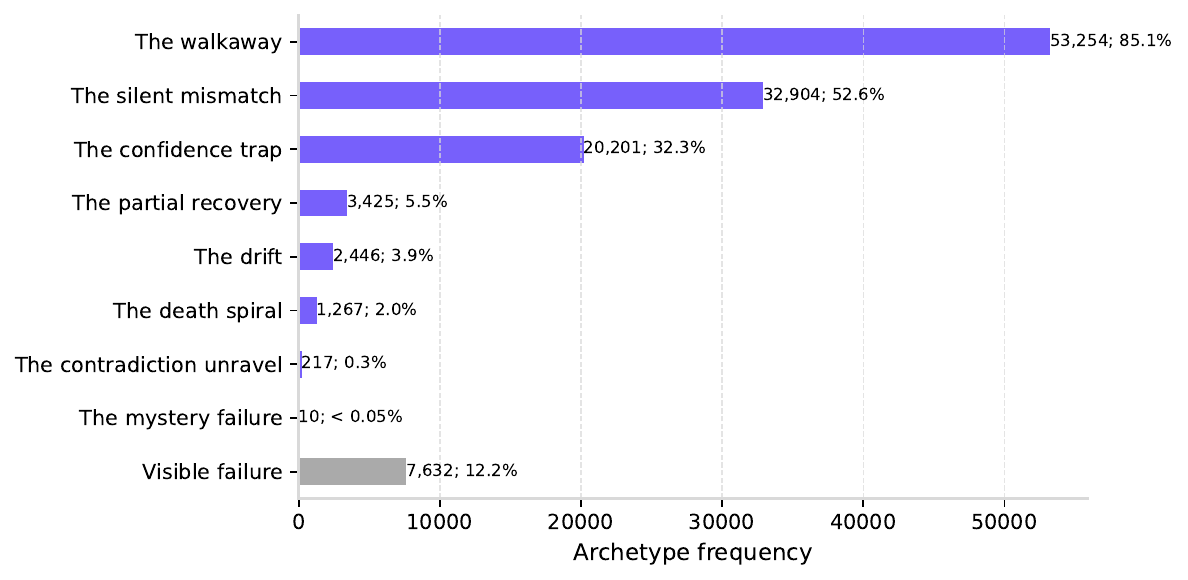}
    \caption{Invisible failure archetype distribution. The bar sizes and counts indicate the frequency of the archetype. Since individual transcripts can manifest multiple archetypes, the percentages are the percent of failure transcripts labeled with that archetype.}
    \label{fig:archdist}
\end{figure}

In our dataset, \stat{62,557} of the \stat{100,000} transcripts (\stat{63\%}) are tagged as involving some kind of failure. Of these cases, \stat{7,632} (\stat{12\%}) are tagged \failtag{visible}, \stat{49,368} (\stat{79\%}) are tagged \failtag{invisible}, and \stat{5,557} (\stat{9\%}) are tagged \failtag{mixed}. This is in itself a significant finding; if a monitoring system relies on detecting overt signals of failure -- e.g., negative sentiment, complaints, explicit correction -- it will catch just \stat{12\%} of failures. 

\Cref{fig:archdist} gives the overall distribution of failures. The top three most frequent archetypes each individually out-number the \arch{Visible failure} cases, which further emphasizes the importance of studying invisible failures.\footnote{There are \stat{7,632} transcripts in our dataset that were marked as \failtag{visible} failures but still received one or more archetype tags. \stat{30\%} are \arch{The partial recovery}, and the others are likely edge cases for the  \failtag{visible}/\failtag{mixed} distinction. We resolve these in favor of the failure mode tags, treating all of them as \arch{Visible failure}, as a precaution against artificially inflating the invisible failure rate.}

\arch{The walkaway} is the most frequent archetype. Its prevalence likely traces to how frequently failures by the AI lead users to abandon the interaction abruptly.
\arch{The silent mismatch} is also extremely common; these are cases in which the AI addresses a different goal than the user intended, but the response is plausible enough that neither party flags the disconnect. \arch{The confidence trap} is a counterpart to this archetype. Here, the AI gives a wrong answer with complete confidence, and the user accepts it. This type of failure is especially insidious because it looks so much like a success; very often, the AI anchors its fabrications in real-sounding sources and uses specificity as a proxy for certainty.

\arch{The mystery failure} is our catch-all category for failures that have no accompanying overt or implicit signals. This archetype appears on only \stat{10} transcripts, suggesting that we have relatively few total blind spots when it identifying failure archetypes.

In the majority of WildChat transcripts, the user has just one turn. In our sample, \stat{63\%} of transcripts are single-turn; \cref{appendix:turns} gives the full distribution. One might wonder whether such interactions account for most of the invisible failures, in particular for \arch{The walkaway}. \Cref{appendix:multiturn} shows that this is not the case; though the percentage of invisible failures drops to \stat{50\%} when we restrict to multi-turn interactions, and \arch{The walkaway} still appears on over \stat{65\%} of interactions. We emphasize too that the quality of single-turn interactions is incredibly important given their prevalence.
     
There are also informative patterns of co-occurrence between the archetypes. To study these, we created a matrix of co-occurrences between pairs of archetypes and reweighted it using two approaches: positive pointwise mutual information (PPMI; \citealt{church-hanks-1990-word,Bullinaria2007}) and conditional distributions (row- or equivalently column-wise normalization). 
PPMI reveals which archetypes co-occur more often than we would expect given their respective frequencies, but it can exaggerate infrequent events, whereas conditional distributions are shaped more by raw frequency. These matrices are given in \cref{appendix:arch-cooccur}. The probability distributions mostly show again how pervasive \arch{The walkaway} is. The PPMI matrix is more illuminating. For example, \arch{The confidence trap} and \arch{The contradiction unravel} are tightly associated, pointing to a pattern of confident but overlooked contradictions from the AI. 
Similarly, \arch{The death spiral} and \arch{The contradiction unravel} are infrequent archetypes that pattern together more than we would expect by chance; this pattern too shines a light on how disruptive contradictions from the AI can be.

\begin{figure}[t]
    \centering
    \includegraphics[width=0.7\linewidth]{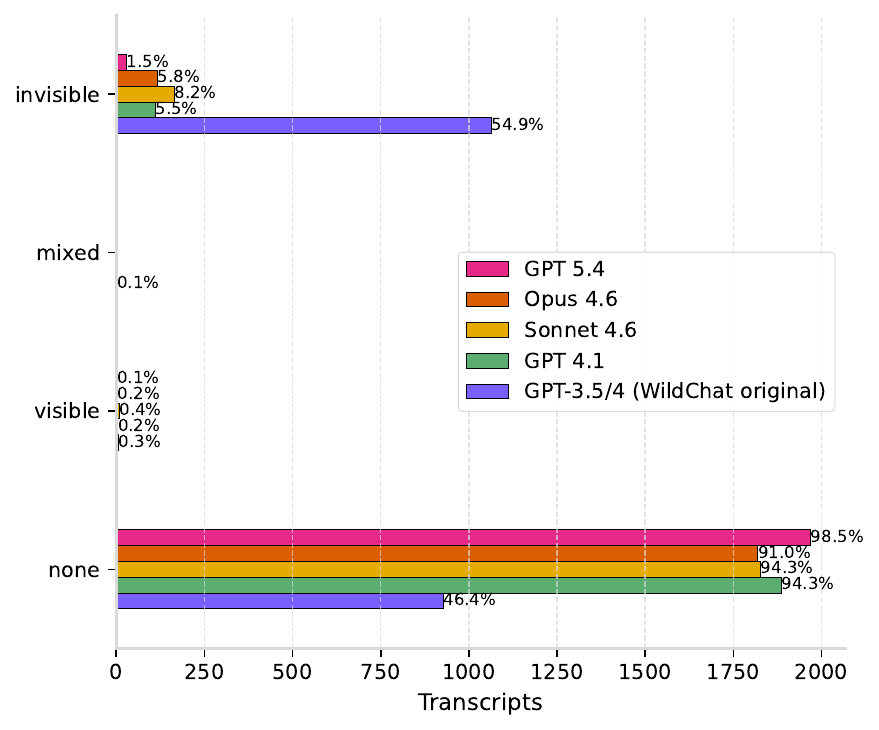}
    \caption{Failure persistence in \ourdataset. The GPT-4 cases are the original WildChat transcripts, included for comparison. Error rates for the newer models are substantially lower (`none' means `no failure detected'), but the vast majority of failures continue to be invisible.}
    \label{fig:failure-persist}
\end{figure}

\section{Estimating invisible failure rates today}\label{sec:future}

The analyses in the preceding section characterize the landscape of invisible failures in WildChat. However, the WildChat transcripts were generated by \mbox{GPT-3.5-Turbo} and \mbox{GPT-4} in 2023--2024. LLMs have improved substantially since then. Are the invisible failure archetypes artifacts of older models, or do they reflect structural dynamics of human--AI interaction that persist across capability levels? We now address this critical question.

\subsection{The \ourdataset\ dataset}\label{sec:methods:validation}

We are not aware of publicly-available data comparable to WildChat that would allow us to systematically study the distribution of invisible failures stemming from today's frontier models; even ShareChat, which is from October 2025, is too old to include \mbox{Opus 4.5} or \mbox{GPT-5.4}. In addition, while there are ongoing data collection efforts for human--AI interactions (most prominently, \citealt{chiang2024chatbotarena}), they generally put the user in the role of (adversarial) evaluator, which leads to very different interaction patterns.

This leads us to adopt a simulation-based approach in which we have current models generate responses to actual user queries from WildChat. This introduces three noteworthy approximations: (1) we have to restrict to single-turn interactions (since we can't assume simulated user responses would have any validity), (2) we have to query the model APIs directly rather than using them as full user experiences (e.g., the difference between GPT-5.4 and ChatGPT), and (3) it doesn't account for how user behaviors have evolved over the past two years. Despite these limitations, we hypothesize that this will provide a meaningful picture of overall progress in the field. In addition, it has the advantage of grounding the experiment directly in the same data that we used for our primary analysis.

Thus, for our experiment, we randomly sampled \stat{2K} single-turn exchanges from the \stat{10K} dataset we used to develop our annotation protocol (\cref{sec:annprocess}). For each sampled user prompt, we generated new responses from 
\stat{Claude Sonnet 4.6, Claude Opus 4.6, GPT-4.1, and GPT-5.4}, to cover a range of recent models and begin to explore how our annotation protocol works on models that are the same, or in the same capability class, as our annotator, \mbox{Opus 4.6}. We adopt a minimal system prompt (``You are a helpful assistant'') and set temperature to 0. We refer to this as `\ourdataset' to distinguish it from our other samples.

\subsection{Invisible failure rates in \ourdataset}\label{sec:failure-persist}

\Cref{fig:failure-persist} summarizes the failure rates for the new models in \ourdataset. Here, we adopt exactly the annotation approach described in \cref{sec:annprocess} and used in our analysis of the \stat{100K} sample. The \mbox{GPT-3.5/4} cases are the actual WildChat interactions in our sample, included here for comparison. Overall, the story is one of progress: failure rates have gone down substantially, from \stat{41.7\%} to  \stat{under 10\%} for the newer models. However, the vast majority of failures continue to be invisible by our standards. 

\begin{figure}[tp]
    \centering
    \includegraphics[width=0.7\linewidth]{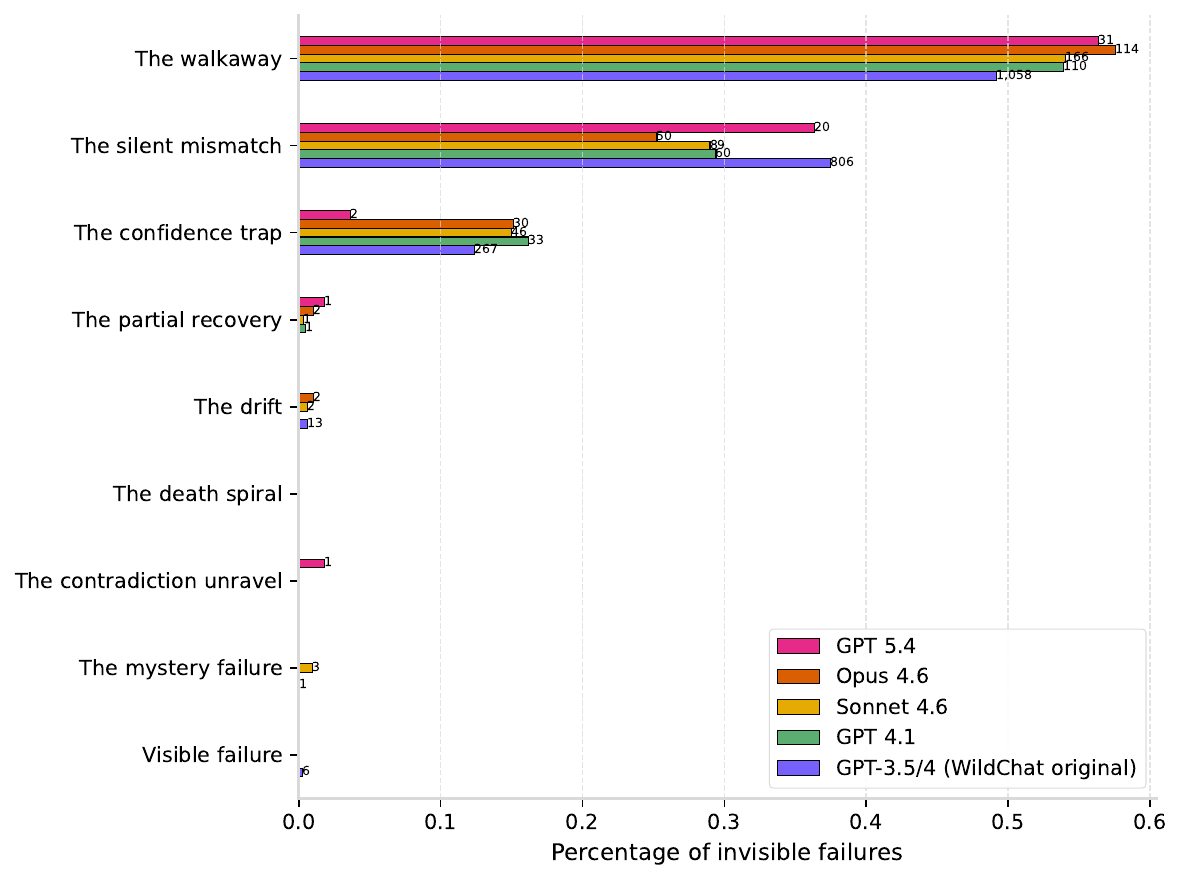}
    \caption{Archetype distribution in \ourdataset. The y-axis retains the frequency ordering of \cref{fig:archdist} for comparison. The overall distribution is quite similar across models; many of the infrequent or unattested archetypes have almost no chance of occurring in the single-turn examples that make up \ourdataset.}
    \label{fig:arch-persist}
\end{figure}

\subsection{Invisible failure archetypes in \ourdataset}

\Cref{fig:arch-persist} shows the distribution of archetypes in \ourdataset, broken out by each model. For this plot, as for \cref{fig:archdist}, we restrict attention to cases marked as invisible failures. To accommodate the variable number of cases marked this way for each model, we normalize the x-axis to be a percentage of this total.

Though the overall rate of failures have gone down (\cref{fig:failure-persist}), the archetype distributions seem to have remained largely stable, with \arch{The walkway}, \arch{The silent mismatch}, and \arch{The confidence trap} still the most prevalent by far. One important caveat here is that the single-turn nature of all the examples in \ourdataset\ means that there are few, if any, chances for \arch{The partial recovery}, \arch{The drift}, \arch{The death spiral}, or \arch{The confidence unravel}. This is why the rates of these archetypes for GPT-3.5/4 (i.e., the actual WildChat transcripts) are lower than in \cref{fig:archdist}. (It is possible that the rates of these have gone \emph{up} in parts of the data that \ourdataset\ cannot cover, though this seems unlikely given the general improvements we see.)

Our two-stage approach to annotation (\cref{sec:annprocess}) allows us to probe the archetype distributions more deeply. In stage~1 of that process, Sonnet, Opus, and GPT-5.4 all provide signal tags, which are then used as the basis for final archetype inference. \Cref{tab:persist-signals} shows the 10 problem signals with the largest change in their distribution between the WildChat original transcripts and the simulations done by Sonnet. The most improved tags are in the top group, and they indicate that models have become better at managing conversational flow and anticipating user needs. The most persistent tags are in the bottom group. The two that have gotten worse seem especially instructive: \signal{over_delivered} may relate to the increasing verbosity of model LLMs, whereas \signal{ai_malfunction} may ultimately trace to users having increasingly broad expectations for what AI will be able to do for them; models improvements may be out-paced by user demands.

\begin{table}[tp]
  \centering  
  \begin{tabular}{l rr rr r r}
    \toprule
     & \multicolumn{2}{c}{Original} & \multicolumn{2}{c}{Sonnet} & $\Delta$ (pp) & \% Eliminated \\
    \cmidrule(lr){2-3} \cmidrule(lr){4-5}
           & $n$ & \%   & $n$ & \%   &        &              \\
    \midrule
    off\_topic\_drift                   &  74 & 3.7 &   0 & 0.0 & $-3.7$ & 100 \\
    plow\_through                       &   7 & 0.4 &   0 & 0.0 & $-0.4$ & 100 \\
    repetition                          &   7 & 0.4 &   0 & 0.0 & $-0.4$ & 100 \\
    performative\_hedge                 &  84 & 4.2 &   3 & 0.2 & $-4.0$ & 96 \\
    conversation\_stalled               & 167 & 8.3 &   8 & 0.4 & $-7.9$ & 95 \\
    \midrule
    generate\_without\_clarifying       & 790 & 39.5 & 434 & 21.7 & $-17.8$ & 45 \\
    silent\_assumption                  & 316 & 15.8 & 199 & 10.0 & $-5.8$ & 37 \\
    ethical\_tension                    & 102 & 5.1 & 106 & 5.3 & $+0.2$ & -4 \\
    over\_delivered                     & 344 & 17.2 & 558 & 27.9 & $+10.7$ & $-$62 \\
    ai\_malfunction                     &  30 & 1.5 &  69 & 3.5 & $+2.0$ & $-$130 \\
    \bottomrule
    \end{tabular}
  \caption{Problem signals most changed between GPT-4 and Sonnet. We restrict attention to signals with at least 5 occurrences in \ourdataset. The top of the table shows the most reduced problem tags, and the bottom shows the most persistent problem tags.}
  \label{tab:persist-signals}
\end{table}

\section{Failure archetypes by domain}\label{sec:domains}

Our annotation effort included a step in which we labeled each transcript for the primary and secondary domains for that interaction. This yields a set of \stat{52} primary domains and \stat{84} secondary domains. For our final analysis, we study the relationship between these domains and our invisible failure archetypes.

The full set of domains is given in \cref{appendix:domains-all}. In \cref{fig:dq-ppmi}, we show the PPMI values derived a matrix in which the rows are domains, the columns are our archetypes, and the cells contain co-occurrence counts. To keep this analysis manageable, we show just the top 10 domains by frequency (though the values are from the full PPMI matrix). This provides a diverse and relevant sample of different goals people have with LLM-based AI: creative writing, user interface design, software development, education, general knowledge, content production, lifestyle, translation, gaming, and IT infrastructure. 

This analysis immediately reveals important contours to the invisible failure landscape. First, though \arch{The walkaway} is the most common archetype (\cref{fig:archdist}), it is not especially overrepresented in any domain. By contrast, \arch{The silent mismatch}, which is also highly frequent, seems to be closely associated with \domain{creative_writing} and, to a lesser extent, \domain{design_ux}. Second, \arch{The confidence trap}, \arch{The contradiction unravel}, and  \arch{The partial recovery} seem to be associated with domains involving factual knowledge and objective capabilities; in these areas, conflicting statements by the AI are likely to be apparent, and this might be key to recovery. Third, and relatedly, the largest value for \arch{Visible failure} is in \domain{software_development}. This is a highly verifiable domain that is dominated by experts who are likely to be able to spot mistakes and inclined to talk back to the AI when it makes mistakes. By contrast, we should perhaps worry about the domains with low rates of \arch{Visible failure}; these are the domains with mistakes that can only be spotted using deep analysis techniques like ours, in part because users are not identifying and overtly complaining about errors.

It should be noted that the PPMI values can exaggerate small values and differences between categories. In \cref{appendix:domain-analyses}, we show a  matrix using the probability of each archetype given each domain and a matrix giving the raw co-occurrence counts. These views of the data do not take the prior frequency of domains or archetypes into account, and they don't directly track differences between observed and expected values the way PPMI values do, but they are a useful counterpoint. In our assessment, they are consistent with all of the observations made above, though the contrasts are weaker, as expected.

\begin{figure}[tp]
    \centering
    \includegraphics[width=1\linewidth]{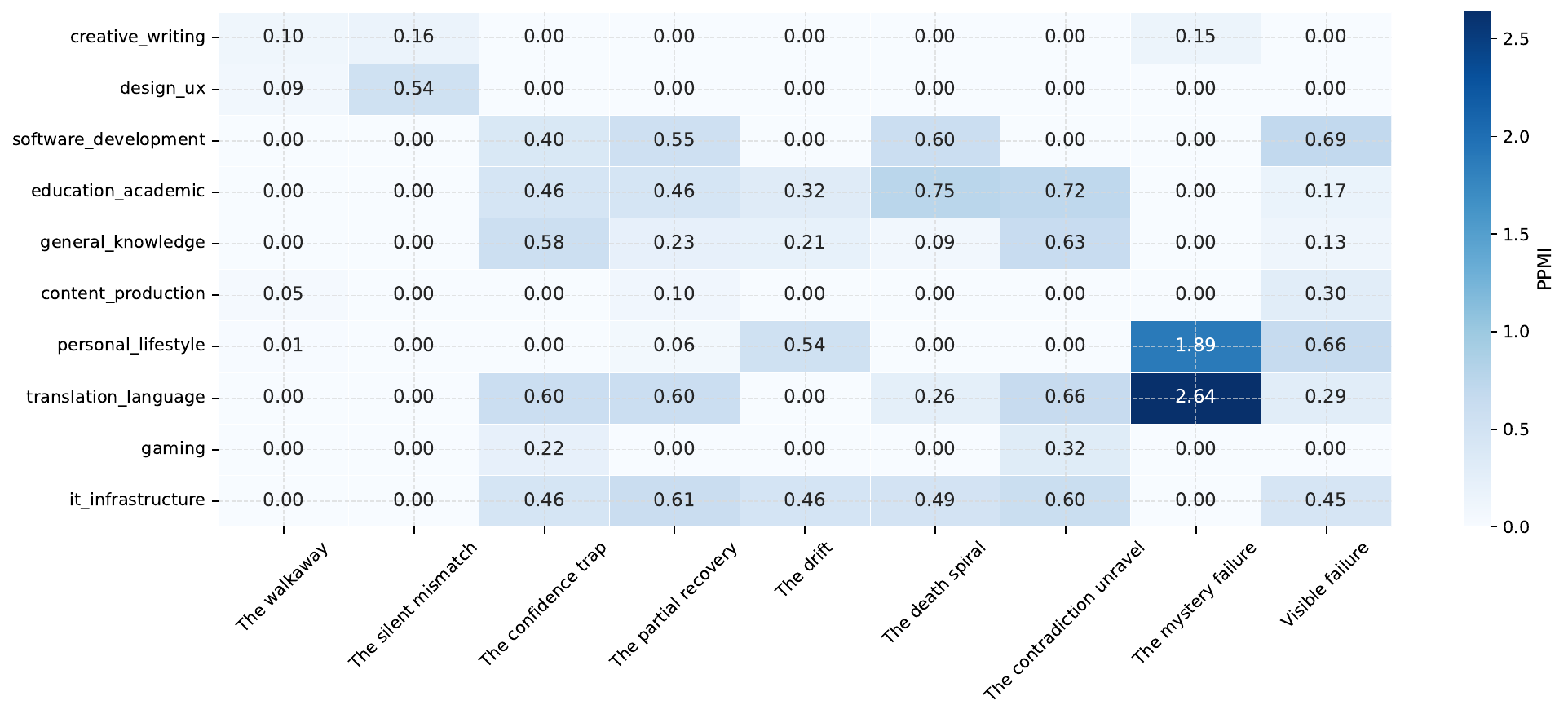}
    \caption{Archetype--domain co-occurrence. The cell values give PPMI values, with darker blues indicating stronger associations. Different domains have different archetype associations, pointing to different underlying challenges for AI systems.}
    \label{fig:dq-ppmi}
\end{figure}

\section{Limitations}\label{sec:limitations}

Our annotation protocol was developed for the English-language subset of WildChat. Applying it to conversations with different characteristics (e.g., voice-based AI, multi-agent systems) would require recalibration. This dataset captures a moment in AI capability, and specific failure rates will shift as models improve. Our retrospective validation (\cref{sec:future}) provides empirical support for the claim that the archetypes are durable. However, this persistence question can only be directly addressed with an ongoing stream of new human--AI interactions, which will be difficult to obtain for open research.

\section{Conclusion}\label{sec:conclusion}

Invisible failures account for \stat{79\%} of all failures in our analysis and cluster into \stat{eight} recognizable archetypes. Our retrospective validation (\cref{sec:future}) provides evidence that these failure patterns are not artifacts of an older model: while failure rates have gone down over the last few years, it remains the case that failures tend to be invisible and manifest one of our invisible failure archetypes.

Because the archetypes are defined in terms of  signal tag reports, they can be computed continuously over any stream of human--AI interactions, making them suitable for monitoring at scale. The fact that different archetypes point toward different interventions -- calibration for \arch{The confidence trap}, alignment-verification mechanisms for \arch{The drift}, interaction-design changes for \arch{The death spiral} -- means the taxonomy supports not just detection but prioritization and action. More broadly, the annotation framework and archetype taxonomy described here can be applied to any corpus of human--AI interactions, enabling teams to surface the invisible failures that existing monitoring approaches miss.

\section*{Reproducibility statement}

The supplementary materials for this submission include all the data and code needed to exactly reproduce all of the analyses in this paper. We also include the annotation scripts required to recreate the annotations we rely on. These, too, should be fully reproducible, up to the variation that still exists even with temperature set to $0$ for current LLM API services.

Our analysis depends on the WildChat dataset. We believe that we have used this dataset in accordance with its license (Open Data Commons Attribution License (ODC-By) v1.0).

\bibliography{colm2026_conference}
\bibliographystyle{colm2026_conference}

\newpage
\appendix

\crefalias{section}{appendix}

\section*{Supplementary materials}

\section{Archetype definitions}\label{appendix:archdefs}

The following are the archetype definitions used in the final prompt to our archetype annotation agent, as implemented in \stat{\texttt{batch_quality_score.py}} in our code repository.

\paragraph{The Confidence Trap} The AI presents incorrect information with unwarranted certainty, and the user accepts it without challenge. The danger is that the interaction looks successful — the AI sounds authoritative, the user seems satisfied — but the user walks away with wrong information. Look for: factual errors delivered without hedging, fabricated specifics, user building on incorrect premises.

\paragraph{The Silent Mismatch} The AI addresses a different goal than the user intended, but the response is plausible enough that neither party flags the disconnect. The AI "answers a question the user didn't ask." Look for: response that is competent but off-target, user's actual need going unaddressed, subtle misinterpretation of the request.

\paragraph{The Drift} The conversation gradually or abruptly loses its connection to the user's original goal. The AI may elaborate on tangentially related topics, add unrequested content, or respond to its own interpretation rather than the user's intent. Can be gradual (verbosity creeping off-topic over turns) or sudden (AI addresses a different but related topic). Look for: response relevance declining over turns, AI addressing adjacent but wrong goals, user's specific requirements getting lost.

\paragraph{The Death Spiral} The conversation enters a repetitive loop. The user keeps asking or correcting, and the AI keeps producing the same or similar output without incorporating the feedback. No progress despite continued effort. Look for: repeated similar responses across turns, user corrections that don't result in changes, escalating user frustration with static AI behavior.

\paragraph{The Contradiction Unravel} The AI contradicts its own prior statements without acknowledging the change. In earlier turns it said X; now it says not-X, with no "I was wrong" or "on reflection." This can erode the user's ability to determine what's actually correct. Look for: incompatible claims across turns, unstated reversals, user potentially confused about which version to trust.

\paragraph{The Walkaway} The conversation ends without resolution and without the user explicitly signaling failure. The user simply stops engaging. This is the hardest archetype to identify because the absence of a signal IS the signal. Look for: unresolved user goal, final AI response that doesn't fully address the need, no subsequent user message, conversation ending without natural closure.

\paragraph{The Partial Recovery} The conversation hits a clear failure but partially recovers. The AI or user identifies the problem and course-corrects, but the recovery is incomplete — the user gets some value but the original goal is not fully met. Look for: error followed by correction, improvement in response quality across turns, but remaining gaps or unaddressed aspects.

\paragraph{The Mystery Failure} The user's goal was not achieved, but no specific failure pattern from the above list explains why. The conversation just... didn't work, and it's hard to point to a specific breakdown. This is a catch-all that flags gaps in our analytical framework. Use sparingly — prefer a specific archetype if one fits even partially.

\newpage

\section{Failure definitions}\label{appendix:failuremode}

The following are the failure-mode definitions used in the final prompt to our archetype annotation agent, as implemented in \stat{\texttt{batch_quality_score.py}} in our code repository.

\paragraph{none} No failure occurred. The conversation succeeded (typically aligns with quality=good, but use your judgment).

\paragraph{visible} The user noticed and reacted to the failure. Look for: explicit corrections, expressions of frustration or dissatisfaction, pointed requests for clarification, the user restating their original request, or direct statements that the AI got something wrong. The user's behavior changed in response to the problem.

\paragraph{invisible} A failure occurred but the user did not catch it or react to it. The conversation may look successful on the surface — the user seems satisfied, doesn't push back, and may even thank the AI — but the AI's output was wrong, off-target, or inadequate in ways the user didn't flag. This is the most dangerous category: the user walks away with a false sense of success.

\paragraph{mixed} The conversation contains BOTH visible and invisible failures. The user caught and reacted to some problems but missed others. For example: the user corrected a factual error (visible) but didn't notice the AI silently dropped one of their requirements (invisible).

\section{Detailed annotation agreement reports}\label{appendix:annotations}

\subsection{Failure annotation agreement reports}\label{appendix:failann}

\begin{table}[H]
\setlength{\tabcolsep}{2pt}
    \begin{subtable}[b]{0.3\textwidth}
    \centering
\begin{tabular}{l cc}
\toprule
 & $\kappa$ & Agree. \\
\toprule
label & kappa & Agreement \\
\midrule
invisible & 0.83 & 0.91 \\
mixed & 0.63 & 0.96 \\
none & 0.85 & 0.93 \\
visible & 0.67 & 0.96 \\
Macro & 0.74 & 0.94 \\
Overall & 0.84 & 0.94 \\
\bottomrule
\end{tabular}
    \caption{Signals-only}
    \end{subtable}
\hfill
    \begin{subtable}[b]{0.3\textwidth}
    \centering
\begin{tabular}{l cc}
\toprule
 & $\kappa$ & Agree. \\
\midrule
invisible & 0.46 & 0.75 \\
mixed & 0.43 & 0.95 \\
none & 0.42 & 0.78 \\
visible & 0.54 & 0.95 \\
Macro & 0.46 & 0.86 \\
Overall & 0.62 & 0.86 \\
\bottomrule
\end{tabular}
    \caption{Transcript-only}
    \end{subtable}
\hfill
    \begin{subtable}[b]{0.3\textwidth}
    \centering
\begin{tabular}{l cc}
\toprule
 & $\kappa$ & Agree. \\
\midrule
invisible & 0.70 & 0.85 \\
mixed & 0.53 & 0.96 \\
none & 0.73 & 0.88 \\
visible & 0.55 & 0.96 \\
Macro & 0.63 & 0.91 \\
Overall & 0.77 & 0.91 \\
\bottomrule
\end{tabular}
    \caption{Signals + Transcript}
    \end{subtable}
    \caption{Failure annotation agreement reports.}
    \label{tab:app:failures}
\end{table}

\subsection{Archetype annotation agreement reports}\label{appendix:archann}

\begin{table}[H]
\setlength{\tabcolsep}{2pt}
    \centering
    \begin{subtable}[b]{0.3\textwidth}
    \centering
\resizebox{1\linewidth}{!}{       
\begin{tabular}{l cc}
\toprule
 & $\kappa$ & Agree. \\
\midrule
The confidence trap & 0.92 & 0.97 \\
The contradiction unravel & 0.54 & 0.99 \\
The death spiral & 0.80 & 0.99 \\
The drift & 0.32 & 0.94 \\
The mystery failure & 0.84 & 0.93 \\
The partial recovery & 0.43 & 0.84 \\
The silent mismatch & 0.73 & 0.89 \\
The visible & 0.00 & 1.00 \\
The walkaway & 0.81 & 0.91 \\
Macro & 0.60 & 0.94 \\
Micro & 0.81 & 0.94 \\
\bottomrule
\end{tabular}
}
    \caption{Signals-only}
    \end{subtable}
\hfill
    \begin{subtable}[b]{0.3\textwidth}
    \centering
\resizebox{1\linewidth}{!}{    
\begin{tabular}{l cc}
\toprule
 & $\kappa$ & Agree. \\
\midrule
The confidence trap & 0.51 & 0.80 \\
The contradiction unravel & 0.49 & 0.99 \\
The death spiral & 0.68 & 0.98 \\
The drift & 0.28 & 0.93 \\
The mystery failure & 0.44 & 0.81 \\
The partial recovery & 0.28 & 0.86 \\
The silent mismatch & 0.50 & 0.75 \\
The visible & 0.00 & 1.00 \\
The walkaway & 0.16 & 0.49 \\
Macro & 0.37 & 0.84 \\
Micro & 0.52 & 0.84 \\
\bottomrule
\end{tabular}
}
    \caption{Transcript-only}
    \end{subtable}
\hfill
    \begin{subtable}[b]{0.3\textwidth}
    \centering
\resizebox{1\linewidth}{!}{       
\begin{tabular}{l cc}
\toprule
 & $\kappa$ & Agree. \\
\midrule
The confidence trap & 0.70 & 0.88 \\
The contradiction unravel & 0.65 & 0.99 \\
The death spiral & 0.77 & 0.98 \\
The drift & 0.32 & 0.93 \\
The mystery failure & 0.73 & 0.89 \\
The partial recovery & 0.28 & 0.83 \\
The silent mismatch & 0.60 & 0.80 \\
The under delivered & 0.00 & 1.00 \\
The walkaway & 0.59 & 0.79 \\
Macro & 0.52 & 0.90 \\
Micro & 0.68 & 0.90 \\
\bottomrule
\end{tabular}
}
    \caption{Signals + Transcript}
    \end{subtable}
    \caption{Archetype annotation agreement reports.}
    \label{tab:app:archann}
\end{table}

\subsection{Signal annotations agreement reports}\label{appendix:signalann}

\begin{table}[H]
    \begin{subtable}[b]{0.48\textwidth}
\resizebox{1\linewidth}{!}{
\begin{tabular}{l cc}
\toprule
 & $\kappa$ & Agree. \\
\midrule
adaptation & 0.71 & 0.94 \\
ai_acknowledges_correction & 0.81 & 0.98 \\
ai_asked_clarifying_question & 0.70 & 0.98 \\
ai_asked_probing_question & 0.59 & 1.00 \\
ai_asks_for_feedback & 0.09 & 1.00 \\
ai_asserts_knowledge_limit & 0.72 & 0.95 \\
ai_cites_source & 0.59 & 0.97 \\
ai_empathy_expressed & 0.38 & 0.99 \\
ai_flags_complexity & 0.60 & 0.96 \\
ai_hedges_uncertainty & 0.57 & 0.92 \\
ai_implicit_refusal & 0.16 & 0.92 \\
ai_malfunction & 0.78 & 0.99 \\
ai_normalizes_difficulty & 0.57 & 0.99 \\
ai_offered_options & 0.41 & 0.97 \\
ai_offers_to_elaborate & 0.48 & 0.96 \\
ai_provides_alternatives & 0.43 & 0.90 \\
ai_provides_caveats & 0.71 & 0.92 \\
ai_provides_example & 0.53 & 0.89 \\
ai_provides_step_by_step & 0.72 & 0.93 \\
ai_references_prior_turn & 0.53 & 0.91 \\
ai_references_user_words & 0.26 & 0.94 \\
ai_refuses_or_declines & 0.76 & 0.98 \\
ai_self_contradiction & 0.10 & 0.98 \\
ai_stated_interpretation & 0.22 & 0.96 \\
ai_structured_response & 0.65 & 0.83 \\
ai_summarizes & 0.37 & 0.94 \\
ai_validates_user & 0.43 & 0.99 \\
ai_warns_user & 0.55 & 0.95 \\
appropriate_confidence & 0.49 & 0.89 \\
appropriate_hedge & 0.35 & 0.90 \\
conversation_advanced & 0.44 & 0.85 \\
conversation_stalled & 0.47 & 0.87 \\
error_commitment & 0.27 & 1.00 \\
error_recovery & 0.59 & 0.99 \\
ethical_tension & 0.50 & 0.95 \\
factual_error & 0.49 & 0.85 \\
false_confidence & 0.46 & 0.81 \\
generate_without_clarifying & 0.21 & 0.63 \\
intent_addressed & 0.47 & 0.84 \\
intent_missed & 0.55 & 0.86 \\
off_topic_drift & 0.42 & 0.94 \\
over_delivered & 0.10 & 0.80 \\
performative_hedge & 0.67 & 0.96 \\
plow_through & 0.35 & 0.97 \\
problem_ignored & 0.56 & 0.85 \\
problem_surfaced & 0.07 & 0.93 \\
repetition & 0.44 & 0.95 \\
scope_matched & 0.37 & 0.76 \\
silent_assumption & 0.20 & 0.81 \\
under_delivered & 0.48 & 0.81 \\[1ex]
Macro & 0.47 & 0.92 \\
Micro & 0.65 & 0.92 \\
\bottomrule
\end{tabular}
}
    \caption{AI signals.}
    \end{subtable}
\hfill
    \begin{subtable}[b]{0.48\textwidth}
\resizebox{1\linewidth}{!}{    
\begin{tabular}{l cc}
\toprule
 & $\kappa$ & Agree. \\
\midrule
user_abandons_thread & 0.72 & 0.96 \\
user_ambiguous_request & 0.52 & 0.81 \\
user_asks_clarification & 0.60 & 0.97 \\
user_corrects_ai & 0.70 & 0.98 \\
user_expresses_dissatisfaction & 0.61 & 0.98 \\
user_expresses_frustration & 0.60 & 0.99 \\
user_implicit_correction & 0.67 & 0.94 \\
user_multi_request & 0.50 & 0.81 \\
user_positive_feedback & 0.81 & 1.00 \\
user_provides_invalid_input & 0.41 & 0.96 \\
user_repeats_request & 0.61 & 0.96 \\
user_scope_change & 0.32 & 0.94 \\
user_validation_seeking & 0.46 & 0.97 \\
Macro & 0.58 & 0.94 \\
Micro & 0.59 & 0.94 \\
\bottomrule
\end{tabular}
}
    \caption{User signals.}
    \end{subtable}
    \caption{Signal annotation agreement reports.}
    \label{tab:app:signals}
\end{table}

\section{Complete domain distribution}\label{appendix:domains-all}

\begin{figure}[H]
    \centering
    \includegraphics[width=0.8\linewidth]{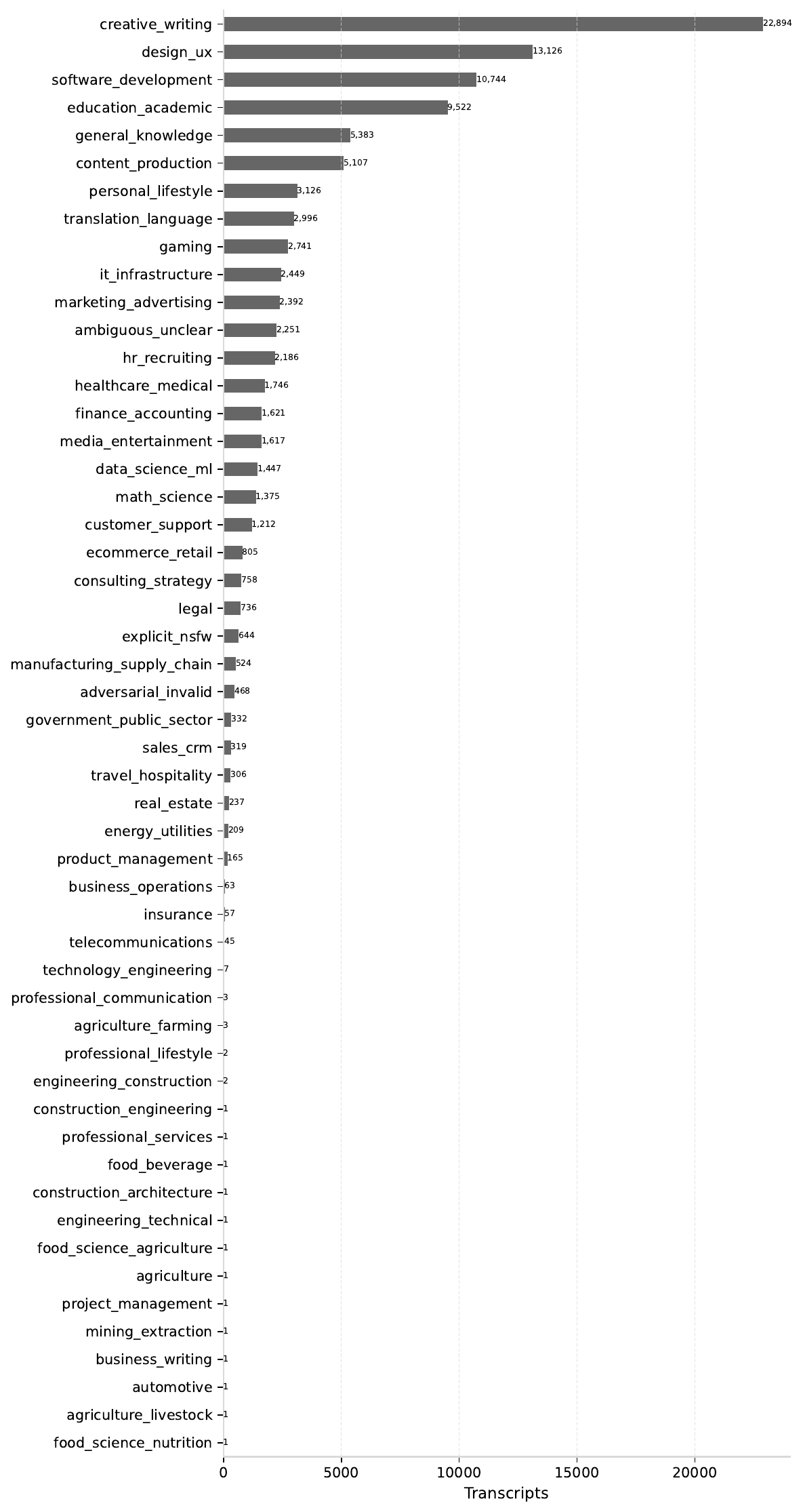}
    \caption{Distribution of domains.}
    \label{fig:domains-all}
\end{figure}

\section{Turn distribution}\label{appendix:turns}

\begin{figure}[H]
    \centering
    \includegraphics[width=1\linewidth]{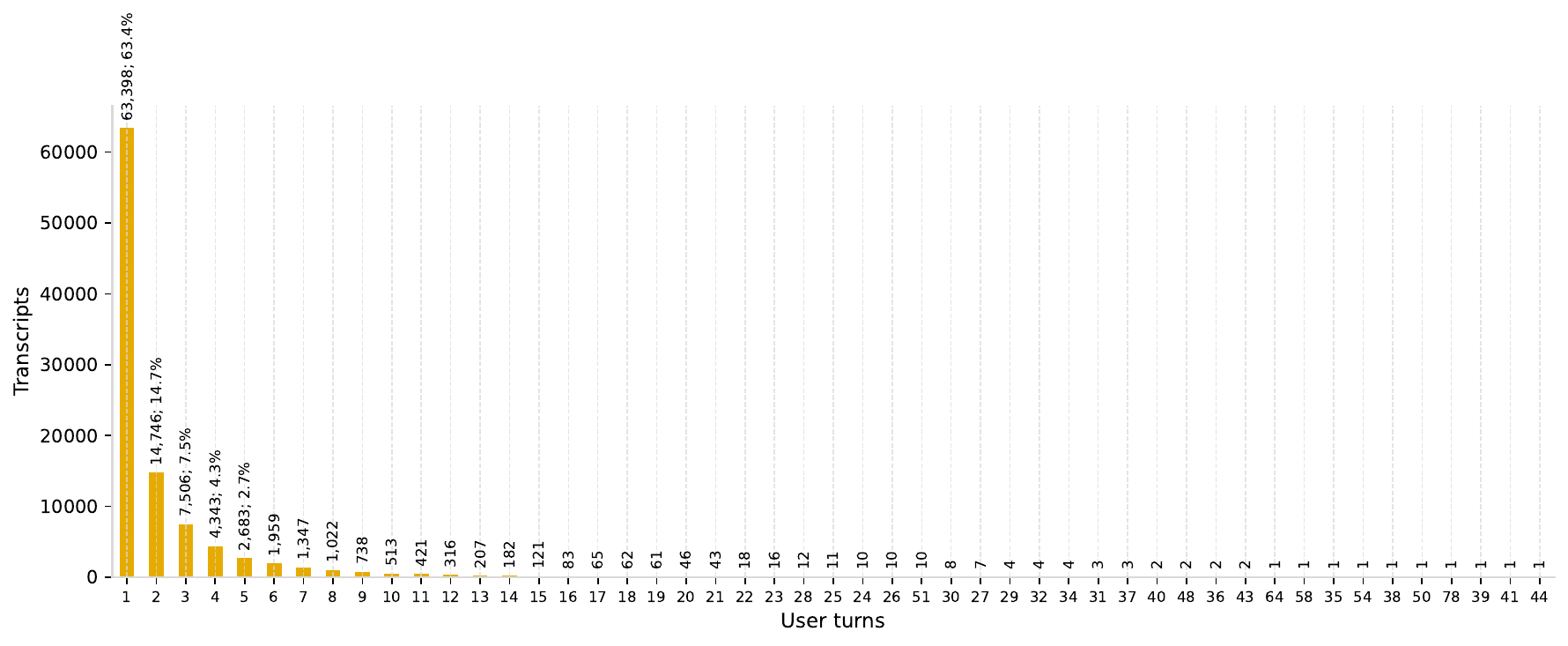}
    \caption{Turn distribution in our sample of WildChat.}
    \label{fig:turns}
\end{figure}

\section{Archetype distribution for multi-turn transcripts}\label{appendix:multiturn}

\begin{figure}[H]
    \centering
    \includegraphics[width=1\linewidth]{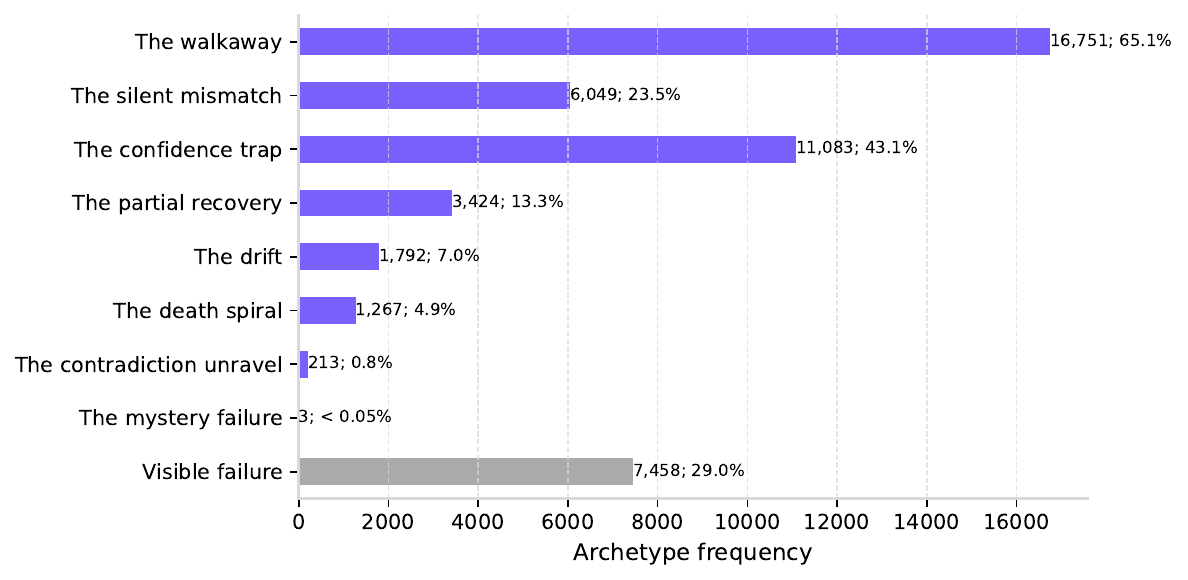}
    \caption{Archetype distribution for multi-turn interactions. The y-axis ordering corresponds to the frequency ordering in \cref{fig:archdist}, to make it easier to track how the prevalence is different in this subset of cases. Though the rate of invisible failures is somewhat lower in this sample, they still account for the majority of failures (\stat{50\%} vs.~\stat{79\%} for the full dataset), and all the archetypes are still attested.}
    \label{fig:app:multiturn:arch}
\end{figure}

\section{Archetype co-occurrence}\label{appendix:arch-cooccur}

\begin{figure}[tp]
    \centering
    \begin{subfigure}[b]{1\textwidth}
        \includegraphics[width=1\linewidth]{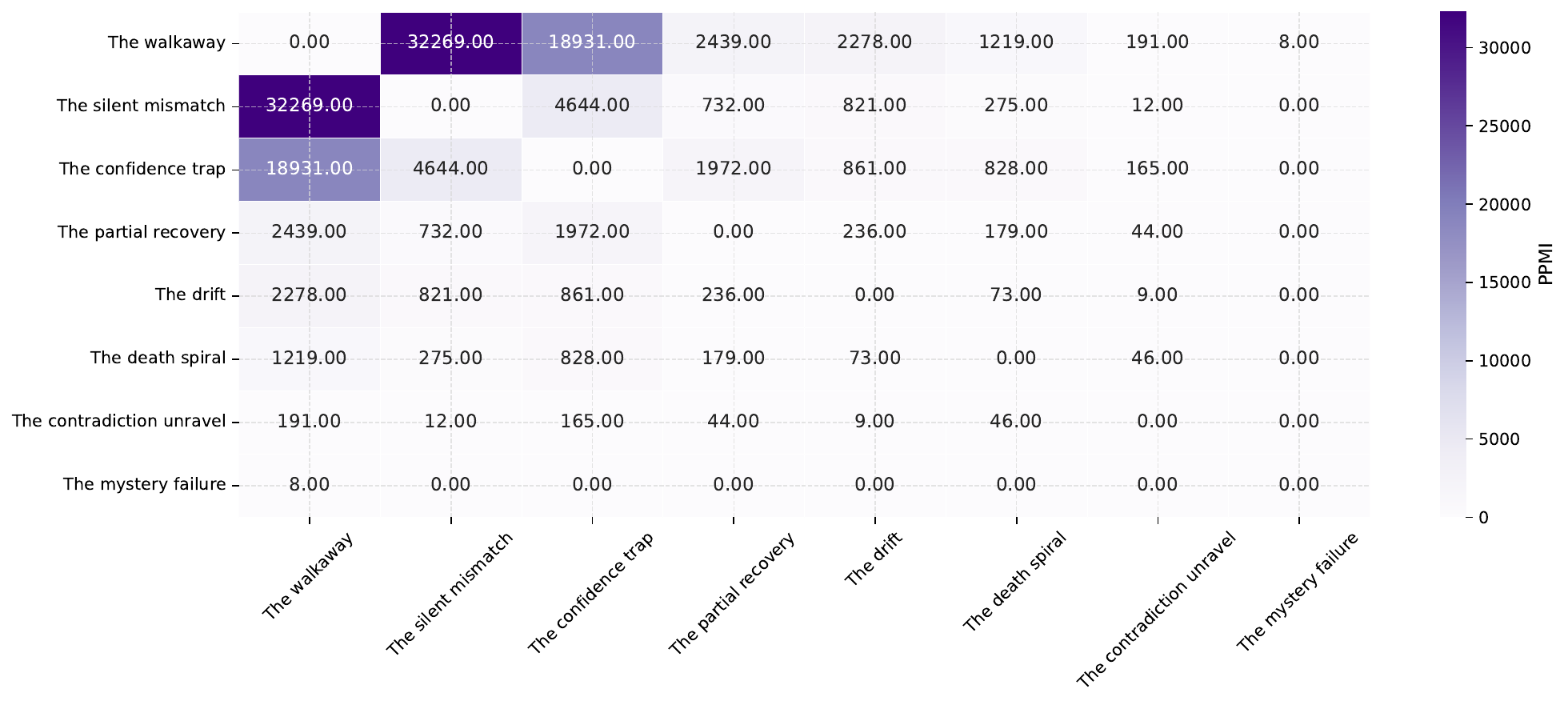}
        \caption{Raw co-occurrence counts.}
    \end{subfigure}

    \begin{subfigure}[b]{1\textwidth}
        \includegraphics[width=1\linewidth]{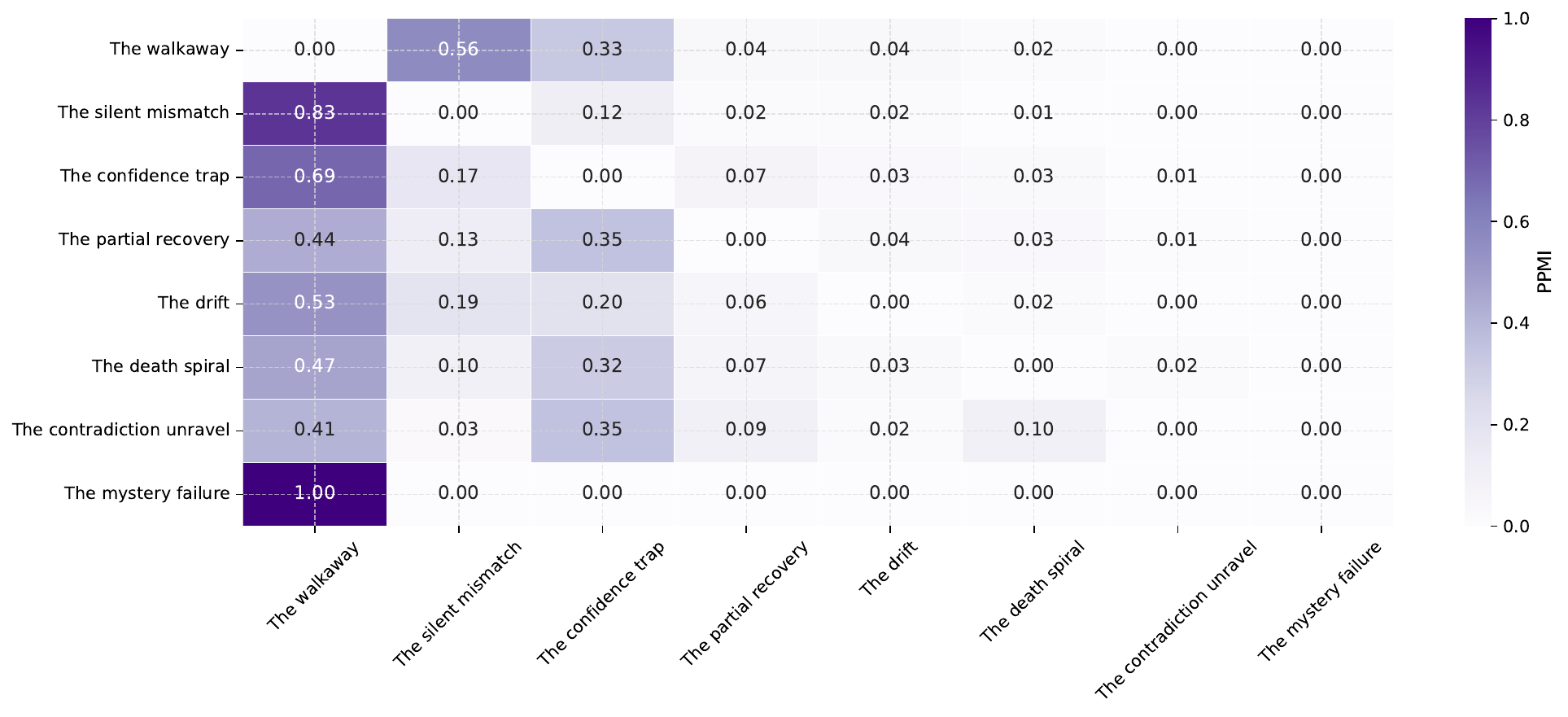}
        \caption{Probability distribution (row-wise normalization).}
    \end{subfigure}
    
    \begin{subfigure}[b]{1\textwidth}
        \includegraphics[width=1\linewidth]{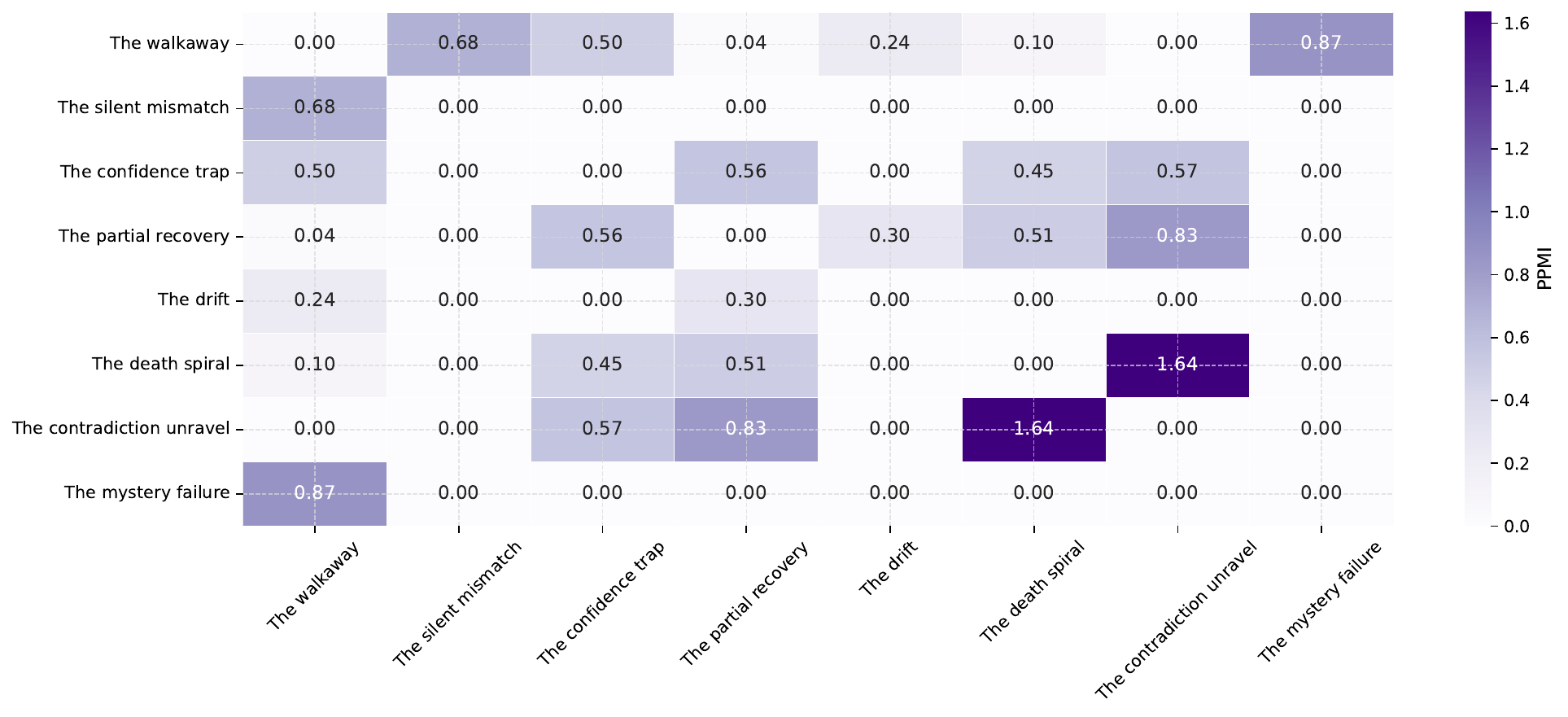}
        \caption{PPMI.}
    \end{subfigure}
    
    \caption{Archetype co-occurrence analyses.}    
    \label{fig:cooccur}
\end{figure}

\Cref{fig:cooccur} provides archetype co-occurrence analyses. The positive pointwise mutual information (PPMI; \citealt{church-hanks-1990-word,Bullinaria2007}) between two categories is defined as
\[
\text{PPMI}(X, a_{i}, a_{j}) = \max\left(0, \frac{P(X_{ij})}{P(X_{i*})\cdot P(X_{*j})}\right)
\]
where $X$ is the matrix of co-occurrences between pairs of archetypes, $P(X_{ij})$ is the probability of $a_{i}$ and $a_{j}$ occurring together, and $P(X_{*i})$ and
$P(X_{*j})$ are row and column probabilities, respectively.

\section{Additional views on the domain--archetype relationship}\label{appendix:domain-analyses}

\begin{figure}[H]
    \centering
    \begin{subfigure}[t]{1\linewidth}
        \centering
        \includegraphics[width=1\linewidth]{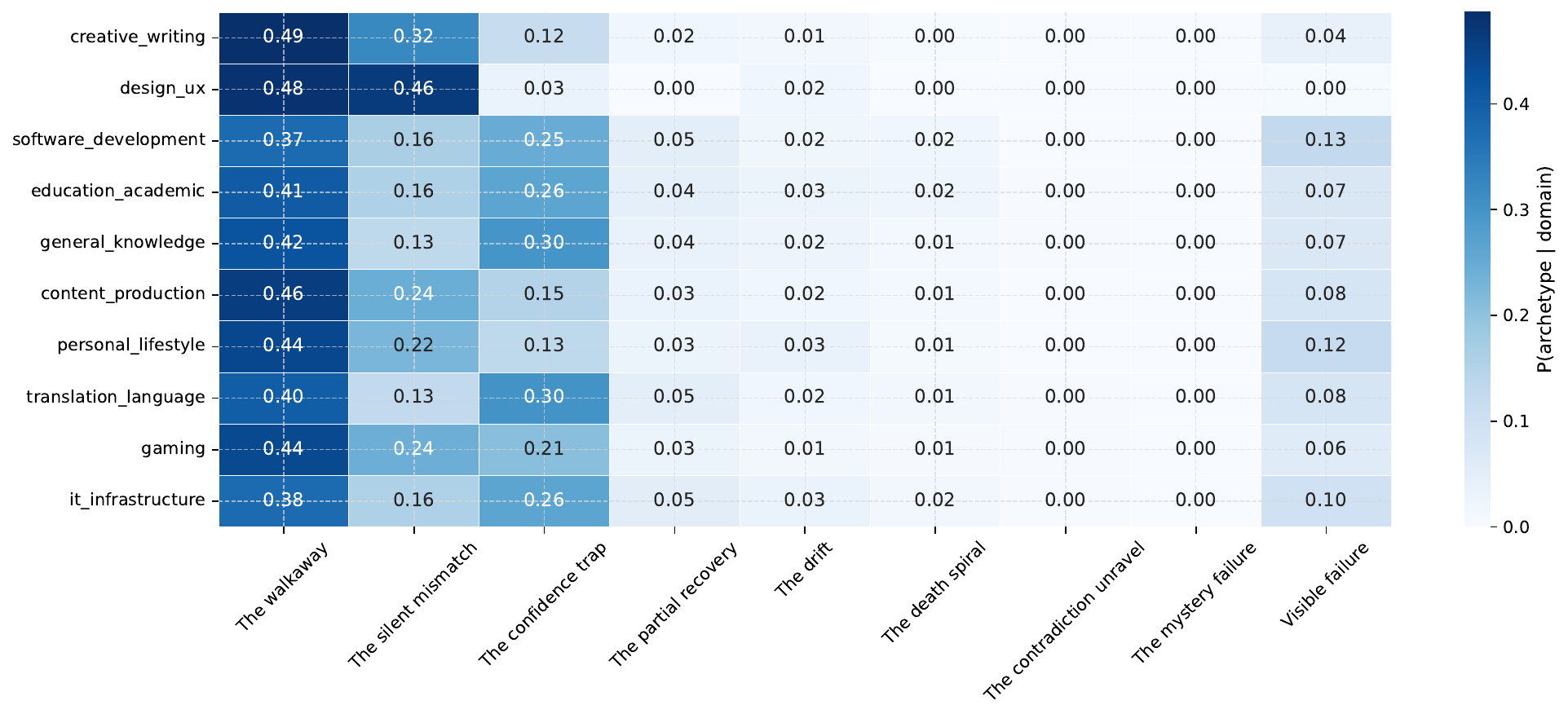}
        \caption{Archetype--domain probabilities.}
        \label{fig:dq-prob}
    \end{subfigure}

    \vspace{10pt}

    \begin{subfigure}[t]{1\linewidth}
        \centering
        \includegraphics[width=1\linewidth]{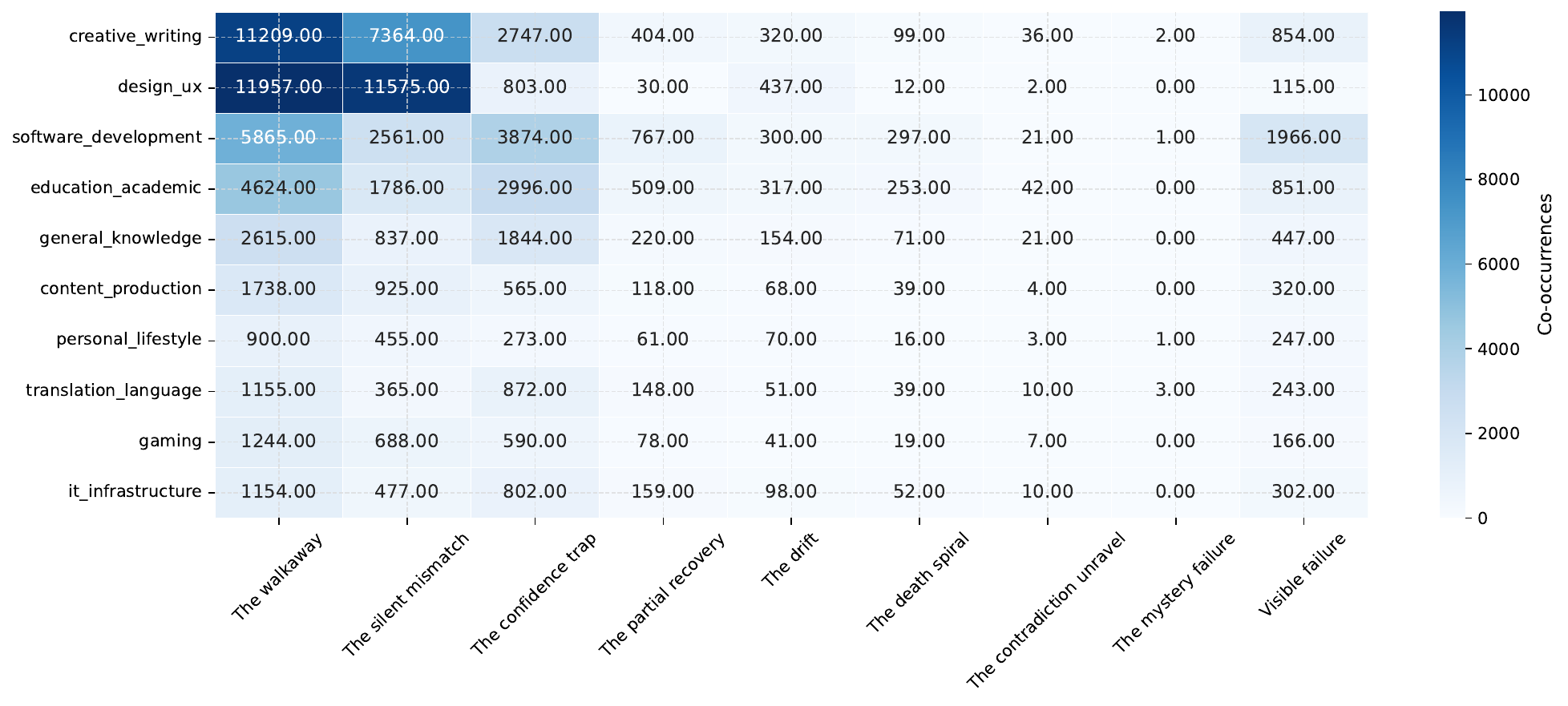}
        \caption{Archetype--domain counts.}
        \label{fig:dq-count}    
    \end{subfigure}
    \caption{Additional views on the domain--archetype relationship.}
\end{figure}

\end{document}